\documentclass[runningheads]{llncs}
\usepackage[T1]{fontenc}
\usepackage{graphicx}
\usepackage[colorlinks,allcolors=blue]{hyperref}
\usepackage{amsmath,amsfonts}
\usepackage{mathtools}

\usepackage{algorithm}%
\usepackage{algorithmicx}%
\usepackage{algpseudocode}%
\usepackage{subfig}
\usepackage{subcaption}
\usepackage[labelfont=bf]{caption}
\usepackage{xcolor}
\usepackage{array}
\usepackage{multirow}
\usepackage[inline]{enumitem}
\usepackage{rotating}
\usepackage{caption}
\usepackage{booktabs}
\usepackage{tikz}
\usepackage{rotating}
\usepackage{tipa} 
\usepackage[nohyperlinks, nolist]{acronym} 
\usepackage[table]{xcolor}
\usepackage{cleveref}
\usepackage{siunitx}
\usepackage{amssymb}
\usepackage{graphicx}
\usepackage{tabularx}
\usepackage{tikz}
\usetikzlibrary{shapes.geometric}

\usepackage{color}

\usepackage[section]{placeins}

\usepackage{amsmath,amsfonts,bm}

\def\eqref#1{equation~\ref{#1}}

\def\1{\bm{1}}

\DeclareMathAlphabet{\mathsfit}{\encodingdefault}{\sfdefault}{m}{sl}
\SetMathAlphabet{\mathsfit}{bold}{\encodingdefault}{\sfdefault}{bx}{n}

\newcommand{\norm}[1]{\left\lVert#1\right\rVert}

\begin{document}
\title{On the Properties of Feature Attribution for Supervised Contrastive Learning\thanks{Accepted at the XAI World Conference 2026.}}
\titlerunning{XAI in Contrastive Learning}

\author{
Leonardo Arrighi\inst{1}\orcidID{0009-0006-2494-0349} \and
Julia Eva Belloni\inst{2,3} \and
Aurélie Gallet\inst{2} \and
Ivan Gentile\inst{4} \and
Matteo Lippi\inst{4} \and
Marco Zullich\inst{5,2}\orcidID{0000-0002-9920-9095}
}
\institute{
University of Trieste, Trieste, Italy \email{leonardo.arrighi@phd.units.it} \and
University of Groningen, Groningen, the Netherlands \and
University of Amsterdam, Amsterdam, the Netherlands \and
IFAB Foundation, Bologna, Italy \and
Delft University of Technology, Delft, the Netherlands \email{m.zullich@tudelft.nl} 
}
\authorrunning{Arrighi et al.}
\maketitle 

\begin{abstract}
Most Neural Networks (NNs) for classification are trained using Cross-Entropy as a loss function.
This approach requires the model to have an explicit classification layer.
However, there exist alternative approaches, such as Contrastive Learning (CL).
Instead of explicitly operating a classification, CL has the NN produce an embedding space where projections of \emph{similar} data are pulled together, while projections of \emph{dissimilar} data are pushed apart.
In the case of Supervised CL (SCL), labels are adopted as similarity criteria, thus creating an embedding space where the projected data points are \emph{well-clustered}.
SCL provides crucial advantages over CE with regard to adversarial robustness and out-of-distribution detection, thus making it a more natural choice in safety-critical scenarios.
In the present paper, we empirically show that NNs for image classification trained with SCL present higher-quality feature attribution explanations than CL with regard to \emph{faithfulness}, \emph{complexity}, and \emph{continuity}.
These results reinforce previous findings about CL-based approaches when targeting more trustworthy and transparent NNs and can guide practitioners in the selection of training objectives targeting not only accuracy, but also transparency of the models.
\end{abstract}



\begin{acronym}
\acro{ai}[AI]{artificial intelligence}
\acro{ann}[ANN]{artificial neural network}
\acro{cnn}[CNN]{convolutional neural network}
\acro{dnn}[DNN]{deep neural network}
\acro{ce}[CE]{Cross-Entropy}
\acro{cl}[CL]{Contrastive Learning}
\acro{fa}[FA]{Feature Attribution}
\acro{scl}[SCL]{Supervised Contrastive Loss}
\acro{tl}[TL]{Triplet Loss}
\acroplural{dnn}[DNNs]{deep neural networks}
\acro{mlp}[MLP]{multi-layer perceptron}
\acro{rnn}[RNN]{recurrent neural network}
\acro{xai}[XAI]{Explainable Artificial Intelligence}
\acro{xuq}[XUQ]{Explainable Uncertainty Quantification}
\acro{uq}[UQ]{Uncertainty Quantification}
\acro{mse}[MSE]{Mean Squared Error}
\acro{mae}[MAE]{Mean Absolute Error}
\acro{rmse}[RMSE]{Root Mean Squared Error}
\acro{ood}[OoD]{out-of-distribution}
\acro{svd}[SVD]{Singular Value Decomposition}
\acro{ssim}[SSIM]{Structural Similarity index}
\acro{aoc}[AoC]{Area over the Curve}
\acro{auc}[AUC]{Area Under the Curve}
\acro{pg}[PG]{Pointing Game}
\acro{al}[AL]{Attribution Localization}
\acro{cm}[CM]{Complexity Metric}
\acro{sgd}[SGD]{Stochastic Gradient Descent}
\acro{pf}[PF]{Pixel Flipping}

\end{acronym}

\section{Introduction}

For \acp{ann} for multi-class classification, the \emph{de facto} framework is to train the model using the \ac{ce} loss function.
However, \acp{ann} trained with \ac{ce} tend to exhibit noticeable undesirable properties, such as
\begin{enumerate*}[label=(\alph*)]
    \item overconfidence in their predictions, which leads to poor \ac{ood} detection \cite{nguyen2015deep}, and
    \item poor robustness to input perturbation, exemplified by the existence of \emph{adversarial attacks} \cite{szegedy2013intriguing}.
\end{enumerate*}
These effects are usually tackled with solutions, e.g., adversarial training \cite{qian2022survey} for the latter, and ensembling \cite{ovadia2019can} for the former.
However, these techniques often come with a large increase in computational load for training and/or inference.
The deployment of \acp{ann} for safety-critical applications, which require robust and reliable models, is hampered by the low sustainability of these approaches. 
\ac{cl} \cite{chopra2005learning} is another approach which relies instead on training a model to produce as output an \emph{embedding space}.
During training, a \emph{similarity} metric between data points is employed, so that projections in this space of similar data points are pulled together, while dissimilar data points are pushed apart.
This methodology can be applied both to an unsupervised or supervised setting: in this case, the similarity criterion is represented by the labels already present in the dataset---data points from the same class will be projected close to each other, while data points from different classes will be projected far apart.
If compared to \ac{ce}, supervised \ac{cl} has shown to produce models with a plethora of desirable properties, including
\begin{enumerate*}[label=(\alph*)]
    \item better clustering in the embedding space if compared to \ac{ce} \cite{khosla2020supervised},
    \item better generalization capabilities \cite{chen2021intriguing,graf2021dissecting},
    \item better adversarial robustness \cite{khosla2020supervised,fan2021does}, and
    \item better uncertainty quantification and \ac{ood} detection capabilities \cite{van2020uncertainty,zhou2021contrastive}.
\end{enumerate*}
This is often achieved with a comparable training (and inference) budget, marking \ac{cl} as a competitive approach to \ac{ce}-trained \acp{ann}, especially in situations where model trustworthiness (e.g., robustness, uncertainty quantification...) is a desirable property.

One of the other areas of trustworthy machine learning that has gained traction in the last decade is \ac{xai}.
\ac{xai} aims at enhancing human oversight over artificial intelligence models by making components of these models more understandable to humans \cite{longo2024explainable}.
\ac{xai} tools can either be local---they provide information which is valid only for one data point or its neighborhood---or global---the information provided is valid for any datapoint, regardless of the region of the data space it belongs to.
In classification tasks, the most common type of XAI tool is \ac{fa} \cite{saarela2024recent}, which aims at ranking the most important features in a model depending on how \emph{salient} they are in determining one or more model predictions.
\ac{fa} explanations are also called \emph{feature importances}, or, in the specific field of image processing \emph{saliency maps} or \emph{heatmaps}.
Especially in the context of models whose predictive dynamics are opaque, \ac{fa} can provide humans with increased intellectual oversight on the models, by giving an indication of which input variables played a role in a given prediction---information which would otherwise be unattainable, e.g., in \acp{ann}.
In these cases, \ac{fa} is usually applied locally, superimposing the saliency values on one single data point.
This can help with, e.g., model debugging \cite{kaur2020interpreting} and identification of hidden biases in the data \cite{zech2018variable,anders2022finding}, albeit Deck et al.~\cite{deck2023critical} posit that this claim is often overstated.
However, in order to be used reliably for these tasks, explanations need to abide to an extensive set of desirable properties which define their quality \cite{nauta2023anecdotal}.
These properties range from the accuracy of representation of the component of the model they explain (\emph{faithfulness}) to how robust these explanations are to small perturbations of the data (\emph{continuity}) to how \emph{interpretable} they are to humans.
In the field of \acp{ann} for imaging, one of the most common \ac{fa} tools is Grad-CAM \cite{selvaraju2020grad}.
It is a local XAI tool originally designed for \acp{cnn} which combines information provided by late-stage activations and gradients to produce visual heatmaps that can be superimposed on the input data.

In the present work, we conducted a comparison in the quality of Grad--CAM-based \acp{fa} generated on \acp{cnn} trained on two popular image datasets---CIFAR10 \cite{krizhevsky2009learning} and Imagenet-S$_{50}$ \cite{gao2022luss}---using \ac{ce} and two loss functions implementing the supervised \ac{cl} paradigm, \ac{scl} \cite{khosla2020supervised} and \ac{tl} \cite{schroff2015facenet}.
We tested our explanations on faithfulness, continuity, contrastivity (how much \ac{fa} changes for different classes), coherence (how much the explanations align with human-generated ground truths), and complexity (how \emph{large} the explanation is).
Our intuition is that, considering the desirable properties of the embedding space produced using \ac{cl} approaches, the quality of these explanations will increase, especially with regard to faithfulness, continuity, and contrastivity.

Our results suggest that \ac{cl}-trained models tend to produce more faithful, continuous, and less complex \acp{fa} than their \ac{ce} equivalents, while having an overall lower contrastivity, and their models achieve good levels of accuracy.
The results on coherence were inconclusive.
Overall, this reinforces the aforementioned findings on the desirable properties connected to the reliability of \ac{cl}, providing further indications that this paradigm should possibly be preferred to \ac{ce} when trustworthiness is required.

The contributions of the present paper are two-fold:
\begin{itemize}
    \item We provide a framework for generating class-specific \ac{fa} explanations in the context of supervised \ac{cl}.
    \item We functionally assess these explanations using state-of-the-art \ac{xai} evaluation metrics, providing evidence of the higher reliability of the explanations generated in the context of supervised \ac{cl}.
\end{itemize}
We believe our findings can be of great use in the field of trustworthy artificial intelligence, by providing indications to boost the reliability of \acp{ann} with little-to-none computing overhead, if compared to more well-established methods.

The code for reproducing our experiments is available at \url{https://github.com/ivan-gentile/CLXAI}.

\section{Related work}\label{sec:related}

\paragraph{\ac{cl}} has emerged as an important paradigm in self-supervised learning, enabling models to learn robust feature spaces without semantic labels. The core principle involves pulling together an ``anchor'' and a ``positive'' sample (typically an augmented view of the same image) while pushing away ``negative'' samples (unrelated images) in the embedding space. Seminal frameworks such as SimCLR \cite{chen2020simple} and MoCo \cite{he2020momentum} demonstrated the effectiveness of \ac{cl} for self-supervised pretraining. However, these methods treat every other image in a dataset as a negative, effectively ignoring semantic class overlap.
To overcome the limitations of instance discrimination, supervised \ac{cl} \cite{khosla2020supervised} extends the contrastive framework to the supervised setting.
Unlike \ac{ce} loss, which focuses on decision boundaries, or self-supervised contrastive loss, which utilizes a single positive pair, the \ac{scl} leverages label information to encourage diverse positives.
It considers \emph{all} samples belonging to the same class as positives, thus pulling them together, while pushing away negative instances.
This formulation creates tighter intra-class clusters and more separable inter-class margins compared to traditional \ac{ce}, leading to improved robustness and generalization \cite{chen2021intriguing,fan2021does}.
We include both \ac{scl} and \ac{tl} due to their different approach to solving the supervised \ac{cl} problem: the former has more of a global structure (``pulling'' and ``pushing'' all instances in the dataset or batch), while the latter takes a more local approach, explicitly considering one positive and one negative sample per anchor.

\paragraph{Local \ac{fa}}
is the task of assigning a score to the features of an \ac{ai} model within the context of a single prediction.
The goal is that of identifying which features were important in determining the prediction, without necessarily eliciting rules that are valid for the whole model.
\ac{fa} has been used for many years as a tool for \emph{variable selection} \cite{guyon2003introduction}; however, the last decade has seen the introduction of several new methods, due to the rise to prominence of \acp{ann}, which, given to their overall lack of transparency, require tools to enhance transparency over their predictive processes.
The 2010s have seen the development of both \emph{model-agnostic} methods---i.e., tools that can be applied to any \ac{ai} model---and \emph{model-specific} methods.
In the former group, LIME \cite{ribeiro2016model} and SHAP \cite{lundberg2017unified} are possibly the most noticeable.
For what concerns \acp{ann}, LRP \cite{bach2015pixel} and DeepLift \cite{shrikumar2017learning} are both widely adopted;
however, for image data processed with \acp{cnn}, the most used tool is certainly Grad-CAM \cite{selvaraju2020grad}.
While the former tools are all operating with raw features (i.e., pixels), Grad-CAM generates explanations using \emph{intermediate features} (e.g., features produced by higher-order convolutional layers).
Ghorbani et al.~\cite{ghorbani2019interpretation}, and Adebayo at al.~\cite{adebayo2018sanity} notice how pixel-level \ac{fa} tools showcase high sensitivity to non-semantic patterns, like high-frequency pixels or generic edges.
Grad-CAM, operating on intermediate features, is less sensitive to these issues, we elect to adopt it as our main \ac{fa} method.
We additionally use a variant of Grad-CAM, called Eigen-CAM \cite{muhammad2020eigen}, as a gradient-free method.
There are other variants of Grad-CAM, for instance, Grad-CAM++ \cite{chattopadhay2018grad}, and Score-CAM \cite{wang2020score} that are widely used in a variety of contexts \cite{he_survey_2023,arrighi_explainable_2023,tang_reviewing_2024,lopes_deep_2022,arrighi_discriminating_2026}.
We decided not to include them in our study either for issues of instability \cite{wilfling2024evaluating} or inefficiency with respect to the two original versions.
{The study which is the closest to the present work is the one by Zhang et al.~\cite{zhang2021excon}, where \ac{fa} is used in the context of self-supervised \ac{cl}.
Their method includes the usage of Grad-CAM within the training loop to enhance the quality of positive pairs.
They report \acp{fa} generated on models trained in this way to have higher faithfulness.
Their analyses did not target supervised learning methodologies, which are usually the \emph{de-facto} solution for classification.}

\paragraph{Assessment of explanations}
All the explanation methods presented so far operate on assumptions that simplify, somehow, the predictive dynamics of the model.
For example, \ac{fa} generally provides feature-level importance scores, ignoring any type of feature interaction, while \acp{ann} are able to learn arbitrarily high degrees of interactions for producing accurate predictions.
Grad-CAM and Eigen-CAM work purely on heuristics and do not have any solid theoretical backing \cite{sundararajan2017axiomatic,rebuffi2020there}.
The evaluation of explanations is a field still in development, and there is currently an attempt at creating standardized frameworks and benchmarks for identifying what actually constitutes a \emph{good} explanation.
The work by Nauta et al.~\cite{nauta2023anecdotal} constitutes possibly the most important attempt at answering these questions.
They created a framework, which they termed ``Co-12'', of twelve facets on which explanations can be evaluated in a \emph{functional} way.
Some of them, like the aforementioned faithfulness and robustness, refer to the content of the model; others refer to the presentation (i.e., representing the explanation in a way that is human-interpretable and does not overwhelm the user); finally, others refer to the context or the task.
Of these properties, faithfulness (or correctness) is possibly the most important to assess.
It measures the alignment between the explanation and the model predictive dynamics.
For instance, unfaithful \acp{fa} may instill a wrong sense of security in model stakeholders, by creating the illusion that models may be using correct features for a prediction, while they may actually be using noise or spurious features \cite{turpin2023language,jacovi2020towards}.
Jacovi and Goldberg \cite{jacovi2020towards} also indicate how it is important to decouple faithfulness from plausibility of an explanation---a property which Nauta et al.~\cite{nauta2023anecdotal} term \emph{coherence}.
Coherence is also important, since it measures the alignment between \emph{artificially-generated} and human-generated explanations---it measures to what degree humans would normally consider an explanation as \emph{sensible}.
Robustness to input perturbations---also called \emph{continuity}---is a desirable property: an explanation should not change if the input is perturbed; however, in the cases in which the output is sensibly changed, a different explanation is desirable---a property termed \emph{contrastivity}.
The size of the explanation (\emph{complexity}) is also important---for instance, humans may find it difficult to interpret large \acp{fa}.
In the present work, we choose to evaluate the properties of faithfulness, coherence, continuity,  contrastivity, and complexity.
We decided to forgo the assessment of the other seven properties, since they are connected specifically to how the explanation is presented, and thus, they go beyond the scope of our research, or they require additional information not present in the dataset or models we work with.
A special note has to be made on the property of \emph{consistency}---the agreement between explanations generated on different models.
While Nauta et al.~\cite{nauta2023anecdotal} indicate it as a desirable property that different models trained on the same data learn similar features, modern research indicates that this very often does not hold---a phenomenon which has also been termed \emph{Rashomon effect} \cite{paes2023inevitability}.
Due to these findings, we forgo the evaluation of consistency.
A summary of the properties evaluated in our work is later presented in \Cref{tab:metrics}.

\section{Methods}
In this section, we will highlight the main methods employed in this paper: the datasets, the \ac{ann} architectures, the loss functions used to train the models, the \ac{xai} tools, and the metrics used to evaluate the explanations.

\subsection{Datasets}
We make use of two image datasets: CIFAR10 \cite{krizhevsky2009learning} and Imagenet-S$_{50}$ \cite{gao2022luss}.
CIFAR10 is a popular image classification dataset. It contains \num{60000} images across 10 different classes. 
The images have size $32\times 32\times 3$.
Imagenet-S$_{50}$ is a dataset for image classification, a subset of the more famous ImageNet \cite{deng2009imagenet}. 
It contains \num{66865} images across 50 different categories.
The images have size $224\times 224\times 3$.
The reason for using this dataset instead of the full Imagenet is due to the annotations: Imagenet-S$_{50}$ provides, in addition to class labels, high-resolution segmentation maps.
These highlight precisely where the class objects are located, making it possible to evaluate the explanations on coherence, as previously done by Wilfling et al.~\cite{wilfling2024evaluating}.

\subsection{Loss Functions}

In the present work, we train models on the aforementioned datasets using three loss functions, \ac{ce}, \ac{scl}, and \ac{tl}.

\paragraph{\ac{ce}}
is the \emph{de facto} choice for a loss function for multi-class classification problems.
Consider an annotated dataset $\{(x_i, y_i)\}_{i=1}^n$. Let $x_i\in\mathcal{X}$ be the input data and $\mathcal{X}$ the data space.
Let $y_i\in\{0,1\}^p$ be the corresponding annotation, a one-hot encoding of size $p$, $p$ being the number of categories.
Consider then a generic model $f:\mathcal{X}\rightarrow[0,1]^p$.
\ac{ce} is then defined as
\begin{equation*}
    \mathcal{L}_\text{CE} = - \sum_{i=1}^{n} \ell_i = - \sum_{i=1}^{n} y_i^\top \log f(x_i).
\end{equation*}
We highlight with $\ell_i$ the individual contribution to the loss for data point $i$.

\paragraph{\ac{scl} \cite{khosla2020supervised}}
is a loss function used for solving embedding problems---it is not originally designed for classification.
Its goal is to train models to project data points with the same label to similar locations of the embedding space, while points with different labels are projected far apart.
For formalizing \ac{scl}, we need to change the nature of our model $f$.
Instead of the previous output space $[0,1]^p$, we replace it with a generic $\mathbb{R}^k$, with $k\lesseqgtr p$; now $f:\mathcal{X}\rightarrow\mathbb{R}^k$.
$\mathbb{R}^k$ acts as the embedding space.
Conversely to \ac{ce}, where each data point is handled independently of the others, in the \ac{scl}, the individual terms $\ell_i$ consider other data points in the dataset.
We call $A(i)$ the set of indices in the dataset, $i$ excluded: $A(i)\doteq \{1,\dots,n\}\setminus i$.
We also consider the set of \emph{positive} examples for $i$, $P(i)$, the indices of the data points with the same label as $i$: $P(i)=\{j\in A(i): y_j=y_i\}$.
The \ac{scl} is defined as

\begin{equation*}
    \mathcal{L}_\text{SCL} = \sum_{i=1}^{n} \ell_i = - \sum_{i=1}^{n} \frac{1}{|P(i)|} \sum_{j\in P(i)} \log \frac{\exp\left(f(x_i)^\top f(x_j)/\tau\right)}
    {\sum_{a\in A(i)}\exp\left(f(x_i)^\top f(x_a)/\tau\right)},
\end{equation*}
where $\tau\in\mathbb{R}^+$ is a \emph{temperature scaling} hyperparameter.

\paragraph{\ac{tl} \cite{schroff2015facenet}} is a contrastive loss that explicitly considers both positive and \emph{negative} examples in the formulation.
Negative examples are a pair of points whose labels differ.
Each data point in the dataset $x_i$ is paired with a positive example $x^{(i)}_+$ and a negative example $x^{(i)}_-$.
The loss is computed as follows:
\begin{equation}\label{eq:tl}
    \mathcal{L}_\text{TL} = \sum_{i=0}^n\max\left\{0, \norm{f(x_i)-f(x^{(i)}_+)}_2^2 - \norm{f(x_i)-f(x^{(i)}_-)}_2^2 + \alpha \right\},
\end{equation}
where $\alpha\in\mathbb{R}^+$ is a \emph{margin} hyperparameter which forces the model not to collapse all embeddings in the same location of the space.
A fundamental issue with the \ac{tl} is the quality of negative samples: negative pairs can represent an arbitrarily difficult problem for the model to solve during the training phase.
``Easy'' negatives are represented by embeddings that are already far from each other (relative to the positive sample and possibly outside the margin, as in \Cref{eq:tl}) and thus contribute little to none to the learning process.
``Hard'' negatives are those that, instead, are closer to the anchor than the positive sample and the anchor are.
Schroff et al.~\cite{schroff2015facenet} introduce the notion of ``semi-hard'' negatives, i.e., negative points which are within the margin $\alpha$ and close to the anchor, but not as close as the positive is.
They posit semi-hard negatives represent an optimal choice for learning, as the optimizer always has a pair of embeddings to ``pull together'' and another pair to ``push apart''.
They term the search for good negative samples during the training process \emph{semi-hard negative mining}.

\subsection{Models}\label{sec:methods_models}

\begin{figure}[t]
    \centering
    \includegraphics[width=0.7\linewidth, trim={5cm 2.6cm 11cm 0}]{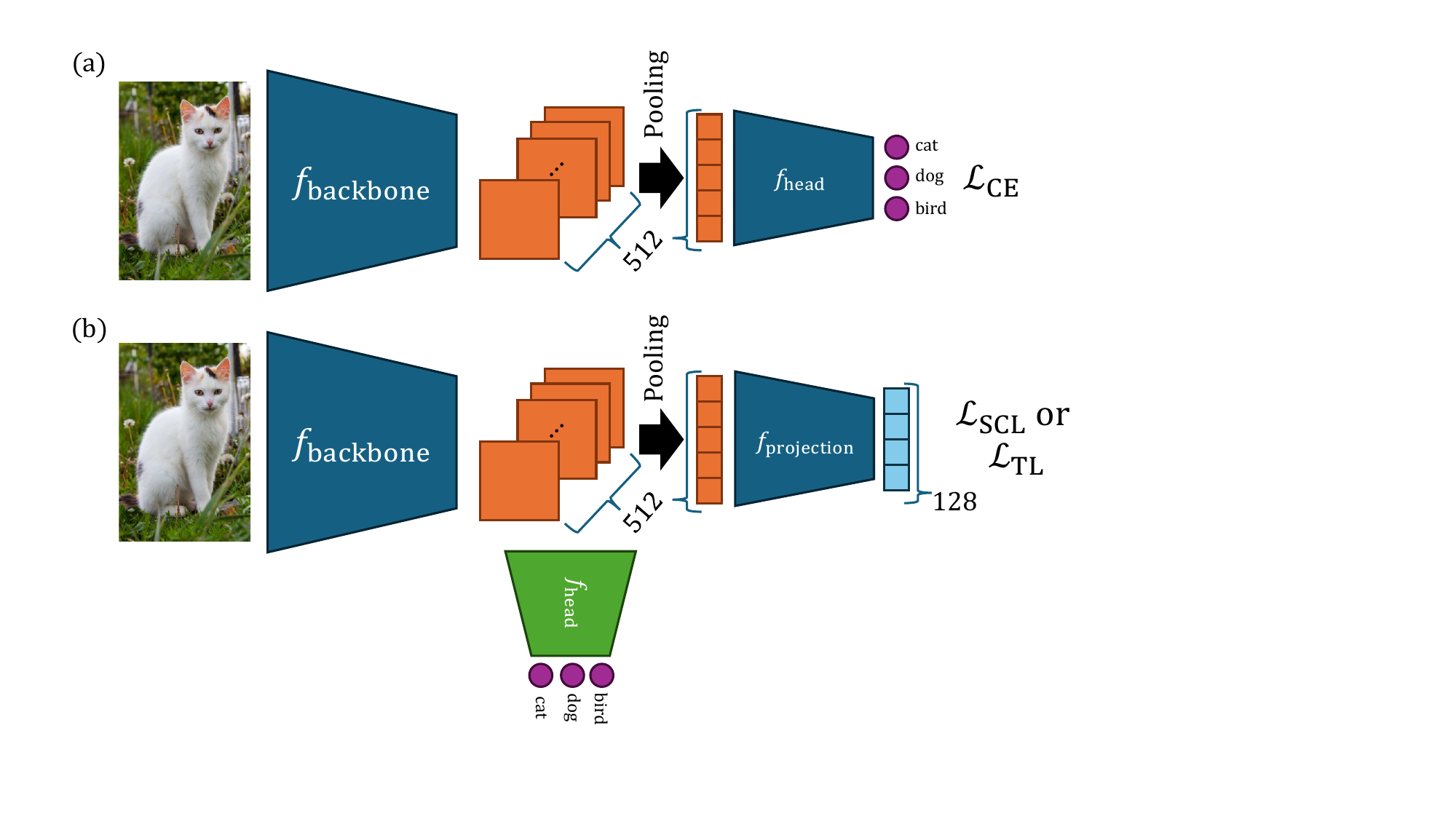}
    
    \caption{Schematization of the two ResNets model variants used.
    (a) represents the classical ResNet, which we train using the \ac{ce} loss ($\mathcal{L}_{\text{CE}}$).
    (b) is the variant we use in \ac{cl}: the classification head $f_\text{head}$ is replaced by a projection head $f_\text{projection}$, which projects the data in $\mathbb{R}^{128}$, where the contrastive losses ($\mathcal{L}_{\text{SCL}}, \mathcal{L}_{\text{TL}}$) are applied. 
    For the sole purpose of the classification, we train a linear classification head $f_\text{head}$, independently from the model, that projects the 512-dimensional embeddings produced by the backbone onto the classification space.
    }
    \label{fig:models_scheme}
\end{figure}

We train two different ResNet \cite{he2016deep} models, a ResNet-18 on CIFAR10 and a ResNet-50 on Imagenet-S$_{50}$.
Both models are \acp{cnn} composed of a \emph{backbone} $f_{\text{backbone}}:\mathcal{X}\rightarrow \mathbb{R}^{h\times w\times 512}$, where the spatial dimensions of the output $h, w$ are dependent upon the size of the input.
The output of the backbone is averaged across spatial dimensions (``Pooling''), and a \emph{classification head}, a simple linear layer with and softmax follows: $f_{\text{head}}:\mathbb{R}^{512}\rightarrow[0,1]^p$.
For the purpose of our work, the initial convolutional layer of the ResNet-18 has a kernel size of $3\times 3$ (to allow for the application of this model to CIFAR10), while, for the ResNet-50, this layer has the regular kernel size of $7\times 7$. 

While this layout is suitable in the case of \ac{ce}, for \ac{cl} we need to change the classification head with a sequence of linear layers of output dimension $k$, so we can train the \acp{cnn} both in the case of \ac{scl} and \ac{tl}. We call this component $f_{\text{projection}}$.
Within our work, we fix $k=128$.
In the context of \ac{cl}, this layout does not immediately allow us to operate the classification.
For this purpose, we train an additional classification head $f_{\text{head}}: \mathbb{R}^{512}\rightarrow\mathbb{R}^p$.
This is the same procedure operated within the self-supervised Sim-CLR \cite{chen2020simple} and reiterated by Khosla et al.~\cite{khosla2020supervised} in the work introducing \ac{scl}: after the training phase, $f_\text{projection}$ is discarded, and the model outputs the embeddings produced by the backbone. 
A schematization of the models is presented in \Cref{fig:models_scheme}.

\subsection{\ac{xai} Tools}

As previously mentioned in \Cref{sec:related}, we make use of two popular \ac{fa} tools, Grad-CAM \cite{selvaraju2020grad}, and Eigen-CAM \cite{muhammad2020eigen}.
Both techniques operate at the level of the feature map $A\in\mathbb{R}^{h\times w\times 512}$ produced by the backbone $f_\text{backbone}$.

\paragraph{Grad-CAM} requires the computation of the gradients of an output logit referred to a specific class $c\in\{1,\dots,p\}$ with respect to the feature map $A$. We call this gradient $g^{(c)}$.
We compute a spatial dimension pooling of $g_c$, obtaining \emph{channel saliencies} $\alpha$:
\begin{equation*}
    \alpha_l = \frac{1}{hw}\sum_{i=1}^{h}\sum_{j=1}^{w} g_{i,j}^{(c)},~~l\in\{1,\dots,512\}.
\end{equation*}
The value $\alpha_l$ assigns an \emph{importance score} to each of the 512 channels of $a$.
The feature map is then multiplied with the importance scores and averaged over the channels to produce the final \ac{fa} over the intermediate features:
\begin{equation*}
    \text{Grad-CAM}_c = \text{ReLU}\left(\frac{1}{512}\sum_{l=1}^{512}\alpha_l\cdot A_{\cdot,\cdot, l}\right),
\end{equation*}
where $A_{\cdot,\cdot, l}$ denotes the activation map for the $l$-th channel.
The ReLU is responsible for removing \emph{negative attributions}, thus producing only attributions that point toward increasing the value of the selected output logit.
To allow overlaying the \ac{fa} matrix on the original image, an upscaling is performed to the original image size.
Finally, the explanation is $[0-1]$ normalized.
Notice that, for each input image, Grad-CAM allows for producing $p$ explanations, one per class.

\paragraph{Eigen-CAM} treats the first principal component of the feature map $A$ as the explanation.
Eigen-CAM first reshapes $A$ to a matrix by combining the two spatial dimensions: $\tilde{A}\in\mathbb{R}^{hw\times 512}$.
Then, it operates \ac{svd} on $\tilde{A}$: $\tilde{A} \overset{\tiny\ac{svd}}{=}U\Sigma V$.
The first column of $U$, $U_{\cdot, 1}\in\mathbb{R}^{hw}$ (the first principal component), is then extracted and converted back to a matrix $\tilde{U}_{\cdot, 1}$, thus producing the desired explanation:
\begin{equation*}
    \text{Eigen-CAM} = \tilde{U}_{\cdot, 1}\in\mathbb{R}^{h\times w}.
\end{equation*}
As in the case of Grad-CAM, upscaling is applied to match the original image size, and $[0-1]$ normalization is applied.
Conversely to Grad-CAM, Eigen-CAM does not allow for generating class-specific explanations---it has been shown that the attributions closely match features connected to the predicted class \cite{muhammad2020eigen}.

\paragraph{Application to \ac{cl}.}
Grad-CAM cannot be applied out-of-the-box to \ac{cl} tasks, since there is no output logit on which to compute the gradients, and the embedded space produced by the models does not have the notion of \emph{output class}.
Similarly to Zhang et al.~\cite{zhang2021excon}, we make use of the additional linear classifier $f_\text{head}$ for generating Grad-CAM explanations.
Since both $f_\text{head}$ and $f_\text{backbone}$ are fully-differentiable, this allows the gradients to flow from output neurons to the feature map $A$, despite these two being different models.
For what concerns Eigen-CAM is concerned, we do not need any adaptation, since it does not consider gradients, allowing it to be freely used within any of our models.
This motivates the inclusion of Eigen-CAM in our study, as a method that generates explanations independently from how the output of the model looks, and can thus be freely applied to any \ac{ann} architecture.

\subsection{Evaluation of \acp{fa}}

\begin{table}[t]
 \caption{Summary of the explanations properties assessed in the present works and corresponding metrics used.
    Notice that we evaluated coherence only for Imagenet-S$_{50}$ (due to the unavailability of segmentation maps on CIFAR10).
    Contrastivity is assessed only for Grad-CAM, since Eigen-CAM does not allow for generating explanations for different classes.
    }
    \centering
    \begin{tabularx}{\textwidth}{p{2.7cm} @{\hspace{0.1cm}} p{4.8cm} @{\hspace{0.2cm}} X}
        \toprule
        \textbf{Property} & \textbf{Description} & \textbf{Metric(s)} \\
        \midrule
        Faithfulness \newline(\emph{correctness}) & Alignment between \ac{fa} and predictive dynamics of model & \ac{pf}\cite{bach2015pixel} \\
        Coherence & Alignment with respect to ground truth (segmentation maps) & \ac{pg} \cite{zhang2018top}, \ac{al} \cite{kohlbrenner2020towards} \\
        Continuity & Robustness to small input perturbations & \ac{ssim} \\
        Contrastivity & Sensitivity to changes in predicted class & \ac{ssim} \\
        Complexity & Size of the explanation & Sparseness \cite{chalasani2020concise}, \ac{cm} \cite{bhatt2020evaluating} \\
        \bottomrule
    \end{tabularx}
    \label{tab:metrics}
\end{table}

We assess the quality of our saliency maps over five facets, all identified by Nauta et al.~\cite{nauta2023anecdotal} in their ``Co-12'' framework: correctness (to which we refer to as \emph{faithfulness}), coherence, continuity, contrastivity, and complexity.

\paragraph{Faithfulness,} the alignment between explanation and predictive dynamics, is evaluated by using variations of the \emph{pixel flipping} approach \cite{adebayo2018sanity}: perturbing important features should cause substantial changes in prediction; conversely, perturbing unimportant features should not change the prediction.
We elect to measure faithfulness by means of \ac{pf} \cite{bach2015pixel}.
\ac{pf} operates by iteratively removing features (pixels, in images) in decreasing order of importance.
For each removal of features, the model's prediction was tracked.
The expectation is that the removal of features should have a decrease in prediction value proportional to the feature importance.
The prediction values are plotted against the proportion of features removed, and faithfulness is computed as the \ac{auc}.
The lower the value, the more faithful the explanation is.
Given an input $x$, we sort its features in descending order according to their importance score.
We denote with $\tilde{x}^{(j)}$ the data point with the $j$-th most important pixels removed.
\ac{pf} is calculated as:

\begin{equation*}
    \operatorname{PF}=\frac{1}{n} \sum_{i=1}^n \operatorname{AUC}\left(f(x_i)_c-f\left(\tilde{x}^{(j)}_i\right)_c\right)_{j=0}^d,
\end{equation*}
where $f(\cdot)_c$ indicates the output value for category $c$.
The lower the \ac{auc}, the more faithful the explanation is.

\paragraph{Coherence} measures the alignment between the explanation and a ground truth representing prior knowledge.
In the present study, we measure coherence as the overlap between the explanations and the high-resolution segmentation maps, which are provided with Imagenet-S$_{50}$.
We suppose, as done by Selvaraju et al.~\cite{selvaraju2020grad}, that, for carrying out the image classification task, the model should mainly concentrate on the object itself rather than on the background or other, semantically uncorrelated objects.
We elect to measure coherence using two metrics:
\begin{enumerate*}[label=(\roman*)]
    \item \ac{pg} \cite{zhang2018top}, and
    \item \ac{al} \cite{kohlbrenner2020towards}.
\end{enumerate*}
\ac{pg} is an accuracy metric which indicates whether the top feature of each explanation lies inside the segmentation maps, while \ac{al} is a precision metric, which indicates how much of the attributions fall inside the segmentation maps.
Considering a segmentation map $s\in\{0,1\}^d$ and the explanation $e\in[0,1]^d$, we compute the metrics as follows:
\begin{equation*}
    \operatorname{PG}=\frac{1}{n}\sum_{i=1}^n s_i\max(e_i),
\end{equation*}
and
\begin{equation*}
    \operatorname{AL}=\frac{1}{n}\sum_{i=1}^n s_i\mathbb{I}[e_i>0].
\end{equation*}
For both metrics, the higher the value, the more coherent the explanation.

\paragraph{Continuity} measures the robustness of the explanation to small input perturbations.
Given a data point $x$, we first generate the \ac{fa} $e$ on this image; we then perturb $x$ with Gaussian noise: $\tilde{x}=x + \epsilon, \epsilon\sim\mathcal{N}(\mathbf{0}, \sigma^2 I)$ and generate the corresponding feature attribution $\tilde{e}$; finally, we measure continuity as the average \ac{ssim} between $e$ and $\tilde{e}$.
Since we desire the two explanations to be similar, the higher the \ac{ssim} score, the better.

\paragraph{Contrastivity} measures the sensitivity of the explanation to changes in the prediction.
For classification tasks, we would like the model to produce different \acp{fa} when explanations are generated on the same data, but different categories.
Considering one data point $x$ and the \ac{ann} output $f(x)_c$, $c$ being the predicted class.
We generate an explanation $e^{(c)}$ for this class.
We then uniformly sample another class from the remaining categories: $c^\prime\sim\{1,...,p\}\setminus c$ and generate the attribution $e^{(c^\prime)}$ for this other category.
We measure contrastivity as average \ac{ssim} between the two explanations \cite{adebayo2018sanity}.
Since we desire the explanations to be different, the higher the norm, the better; the lower the \ac{ssim}, the better.

\paragraph{Complexity} measures how \emph{large} the explanation is.
Larger explanations are usually difficult to interpret for humans; thus, compact \acp{fa} are usually preferred.
In image data, this requirement would translate to saliency maps which are more \emph{focused} rather than wide.
We measure the size by means of
\begin{enumerate*}[label=(\roman*)]
    \item Sparseness, and
    \item \ac{cm} \cite{bhatt2020evaluating}.
\end{enumerate*}
\ac{cm} is measured as the average entropy of the explanation:
\begin{equation*}
    \operatorname{CM} = -\frac{1}{n}\sum_{i=1}^n\sum_{j=1}^d \tilde{e}_{i,j}\log \tilde{e}_{i,j},
\end{equation*}
where $\tilde{e}_i\in\mathbb{R}^{d}$ is a normalized explanation, i.e., such that the attributions are all positive and sum to 1.

\section{Experimental Setup}
\label{sec:experimental_setup}

In this section, we provide all the details necessary for reproducing our experiments, including training configurations, data augmentation strategies, and explanation generation parameters.

\subsection{Training Configuration}

We summarize the training hyperparameters in \Cref{tab:hyperparams}.
All models are trained from scratch using a cosine annealing learning rate schedule.
For \ac{scl}, we adopt the temperature value $\tau = 0.07$ as recommended by Khosla et al.~\cite{khosla2020supervised}.
For \ac{tl}, we use a margin of $0.3$, and we adopt the in-batch semi-hard negative mining strategy illustrated by Hermans et al.~\cite{hermans2017defense}.
The projection head $f_\text{projection}$ consists of two fully-connected layers with ReLU activation, mapping the backbone features to a 128-dimensional embedding space, followed by L2 normalization.

For classification with \ac{scl}-trained models, we freeze the encoder and train a linear classifier $f_\text{head}$ on the learned representations for 100 epochs using \ac{sgd} with learning rate 0.1.

\begin{table}[t]
    \centering
    \caption{Training hyperparameters for all model configurations. 
    For SCL and TL, after contrastive/metric pretraining, we train a linear classifier on frozen features for 100 epochs with a learning rate of 0.1. 
    All models are trained from scratch without pretrained weights.
    TL uses semi-hard negative mining, falling back to hard mining when no semi-hard triplets are available.
    }
    \begin{tabular}{l c c c c c c}
        \toprule
        \textbf{Parameter} & \multicolumn{3}{c}{\textbf{CIFAR10}} & \multicolumn{3}{c}{\textbf{Imagenet-S$_{50}$}} \\
        \cmidrule(lr){2-4} \cmidrule(lr){5-7}
        & CE & SCL & TL & CE & SCL & TL \\
        \midrule
        Architecture & ResNet-18 & ResNet-18 & ResNet-18 & ResNet-50 & ResNet-50 & ResNet-50 \\
        Epochs & 200 & 500 & 500 & 160 & 160 & 160 \\
        Batch size & 128 & 256 & 256 & 512 & 512 & 512 \\
        Optimizer & SGD & SGD & SGD & AdamW & RAdam & RAdam \\
        Learning rate & 0.1 & 0.5 & 0.1 & $10^{-3}$ & $10^{-3}$ & $10^{-3}$ \\
        Momentum & 0.9 & 0.9 & 0.9 & --- & --- & --- \\
        Weight decay & $5\times 10^{-4}$ & $10^{-4}$ & $10^{-4}$ & $10^{-4}$ & $10^{-4}$ & $10^{-4}$ \\
        Temperature $\tau$ & --- & 0.07 & --- & --- & 0.07 & --- \\
        Margin $\alpha$ & --- & --- & 0.3 & --- & --- & 0.3 \\
        Negatives mining & --- & --- & semi-hard & --- & --- & semi-hard \\
        Warmup epochs & --- & 10 & 10 & 5 & 5 & 5 \\
        LR schedule & Cosine & Cosine & Cosine & Cosine & Cosine & Cosine \\
        \bottomrule
    \end{tabular}
    
    \label{tab:hyperparams}
\end{table}

\subsection{Data Augmentation}

\paragraph{CIFAR10.}
For \ac{ce} training, we apply standard augmentations:
\begin{enumerate*}[label=(\roman*)]
    \item random cropping with 4-pixel padding, and
    \item random horizontal flipping with probability 0.5.
\end{enumerate*}
For \ac{scl}, following Chen et al.~\cite{chen2020simple}, we employ stronger augmentations:
\begin{enumerate*}[label=(\roman*)]
    \item random resized cropping (scale 0.2--1.0),
    \item random horizontal flipping,
    \item color jittering (brightness, contrast, saturation $\pm$0.4, hue $\pm$0.1) with probability 0.8, and
    \item random grayscale conversion with probability 0.2.
\end{enumerate*}
All images are normalized using dataset statistics ($\mu=(0.491, 0.482, 0.447)$, $\sigma=(0.247, 0.244, 0.262)$).

\paragraph{Imagenet-S$_{50}$.}
For \ac{ce} training, we use random resized cropping to $224\times224$ pixels and random horizontal flipping.
For \ac{scl}, we additionally apply:
\begin{enumerate*}[label=(\roman*)]
    \item color jittering (brightness, contrast, saturation $\pm$0.8, hue $\pm$0.2) with probability 0.8,
    \item random grayscale with probability 0.2, and
    \item Gaussian blur ($\sigma \in [0.1, 2.0]$) with probability 0.5.
\end{enumerate*}
Standard ImageNet normalization is applied ($\mu=(0.485$, $ 0.456, 0.406)$, $\sigma=(0.229, 0.224, 0.225)$).
At test time, images are resized to 256 pixels on the shorter side and center-cropped to $224\times224$.

\subsection{Explanation Generation and Evaluation}

We generate saliency maps on the test splits of CIFAR10 and Imagenet-S$_{50}$ using the \texttt{pytorch-grad-cam} library \cite{jacobgilpytorchcam}.
For both Grad-CAM and Eigen-CAM, we target the final convolutional block of the backbone (\texttt{layer4[-1]} in PyTorch notation).
For Grad-CAM, explanations are computed with respect to the model's predicted class rather than the ground-truth label, as this better reflects the model's actual decision-making process.
All saliency maps are min-max normalized to $[0, 1]$ before evaluation.

For the evaluation of continuity, we add Gaussian noise with standard deviation $\sigma = 0.02$ to the normalized input images.

\subsection{Implementation Details}

All experiments were conducted using PyTorch 2.1.0.
We use the \texttt{quantus} library \cite{hedstrom2023quantus} for assessing the explanations, except for the \ac{ssim} metric, for which we use the implementation found in \texttt{scikit-image} \cite{walt2024scikitimage} due to its absence from \texttt{quantus}.
All experiments use a fixed random seed for reproducibility.
We train on five different seeds per loss.

\section{Results}

\begin{table}[ht!]
    \centering
    \caption{Accuracy values (mean over five runs $\pm$ std) for the two models trained using \ac{ce} loss, \ac{scl}, and \ac{tl}.
    The results for \ac{scl} and \ac{tl} are computed via the extra linear classification head $f_\text{head}$ applied to the embeddings produced by the models.}
    \begin{tabular}{l c c}
        \toprule
         \textbf{Loss }& {\textbf{CIFAR10}} & {\textbf{Imagenet-S$_{50}$}} \\
        \midrule
         \ac{ce} & $95.32 \pm 0.10$ & $89.17 \pm 0.38$ \\
         \ac{scl} & $94.83 \pm 0.12$ & $85.85 \pm 0.29$ \\
         \ac{tl} & $94.68 \pm 0.17$ & $82.55 \pm 0.64$ \\
        \bottomrule
    \end{tabular}
    
    \label{tab:accuracy}

    \centering
    \caption{
    Summary of the results (mean over five runs $\pm$ std) for the evaluation of the \acp{fa} for every combination of metric, dataset, and loss function.
    Values in \textbf{bold} indicate the best result for the specific metric and \ac{fa} technique.
    Arrows pointing upward $(\uparrow)$ indicate that, for the specific metric, the higher the value, the better.
    Arrows pointing downward $(\downarrow)$ indicate the opposite.
    Contrastivity values on Eigen-CAM are not available because this technique does not allow to generate explanations for different categories.
    For what concerns the coherence scores are concerned, the high variance of the results does not allow us to determine the best score between the various techniques.
    }

    \scalebox{0.8}{
    \begin{tabular}{l c c c c c c  }
        \toprule

         \textbf{CIFAR10}& \multicolumn{3}{c}{\textbf{Grad-CAM}} & \multicolumn{3}{c}{\textbf{Eigen-CAM}} \\
        \cmidrule(lr){2-4} \cmidrule(lr){5-7}
        & \ac{ce} & \ac{scl} & \ac{tl} & \ac{ce} & \ac{scl} & \ac{tl}  \\
        \midrule
        Faithfulness -- \ac{pf} $(\downarrow)$ & $0.37 \pm 0.01$ & $\mathbf{0.27 \pm 0.00}$ & $0.32 \pm 0.01$ & $0.36 \pm 0.01$ & $\mathbf{0.26 \pm 0.00}$ & ${0.32 \pm 0.01}$\\
        Continuity -- \ac{ssim} $(\uparrow)$ & $0.54 \pm 0.02$ & $\mathbf{0.57 \pm 0.01}$ & $0.57 \pm 0.02$ & $0.49 \pm 0.03$ & $0.69 \pm 0.01$ & $\mathbf{0.80 \pm 0.01}$\\
        Contrastivity -- \ac{ssim} $(\downarrow)$ & ${0.003 \pm 0.003}$ & ${0.006 \pm 0.002}$ & $\mathbf{0.001 \pm 0.001}$ & --- & --- & ---\\
        Complexity -- \ac{cm} $(\downarrow)$ & $6.81 \pm 0.00$ & $\mathbf{6.39 \pm 0.01}$ & $6.71 \pm 0.00$ & $6.15 \pm 0.01$ & $\mathbf{6.00 \pm 0.01}$ & ${6.05 \pm 0.01}$\\
        \bottomrule
    \end{tabular}
    }
    
    \par\vspace{1em}
    
    \scalebox{0.8}{
    \begin{tabular}{l c c c c c c  }
        \toprule

         \textbf{Imagenet-S$_{50}$}& \multicolumn{3}{c}{\textbf{Grad-CAM}} & \multicolumn{3}{c}{\textbf{Eigen-CAM}} \\
        \cmidrule(lr){2-4} \cmidrule(lr){5-7}
        & \ac{ce} & \ac{scl} & \ac{tl} & \ac{ce} & \ac{scl} & \ac{tl}  \\
        \midrule
        Faithfulness -- \ac{pf} $(\downarrow)$ & $0.17 \pm 0.00$ & $\mathbf{0.10 \pm 0.00}$ & $0.19 \pm 0.01$ & $0.18 \pm 0.01$ & $\mathbf{0.13 \pm 0.00}$ & ${0.22 \pm 0.01}$\\
        Continuity -- \ac{ssim} $(\uparrow)$ & $0.38 \pm 0.00$ & $\mathbf{0.61 \pm 0.00}$ & $0.39 \pm 0.01$ & $0.61 \pm 0.01$ & $\mathbf{0.66 \pm 0.00}$ & ${0.45 \pm 0.01}$\\
        Contrastivity -- \ac{ssim} $(\downarrow)$ & $\mathbf{0.20 \pm 0.00}$ & ${0.34 \pm 0.00}$ & ${0.28 \pm 0.00}$ & --- & --- & ---\\
        Complexity -- \ac{cm} $(\downarrow)$ & $10.20 \pm 0.01$ & $\mathbf{9.70 \pm 0.01}$ & $10.07 \pm 0.01$ & $9.66 \pm 0.01$ & $\mathbf{9.40 \pm 0.01}$ & ${10.04 \pm 0.02}$\\
        Coherence -- \ac{pg} $(\uparrow)$ & $0.84 \pm 0.37$ & $0.85 \pm 0.36$ & $0.67 \pm 0.47$ & $0.83 \pm 0.37$ & $0.80 \pm 0.40$ & $0.62 \pm 0.49$\\
        Coherence -- \ac{al} $(\uparrow)$ & $0.63 \pm 0.26$ & $0.68 \pm 0.24$ & $0.55 \pm 0.27$ & $0.71 \pm 0.27$ & $0.68 \pm 0.29$ & $0.54 \pm 0.27$\\
        \bottomrule
    \end{tabular}
    }

    \label{tab:xai_metrics}
\end{table}

\begin{figure}[ht!]
    \centering
    \scalebox{0.8}{
    \begin{tikzpicture}
        \node at (0,2.4) {\textbf{Original image}};
        \node at (2.5, 2.4) {\textbf{CE}};
        \node at (4.8, 2.4) {\textbf{SCL}};
        \node at (7.1, 2.4) {\textbf{TL}};
        \node at (9.2, 1.1) {\textbf{Grad-CAM}};
        \node at (9.2, -1.1) {\textbf{Eigen-CAM}};
        \node at (9.2, -3.4) {\textbf{Grad-CAM}};
        \node at (9.2, -5.6) {\textbf{Eigen-CAM}};
        \node at (9.2, -7.9) {\textbf{Grad-CAM}};
        \node at (9.2, -10.1) {\textbf{Eigen-CAM}};
        \node at (9.2, -12.4) {\textbf{Grad-CAM}};
        \node at (9.2, -14.6) {\textbf{Eigen-CAM}};

        \node[rotate=90] at (-1.75, -2.25) {\large\textbf{CIFAR10}};
        
        \node at (0,0) {\includegraphics[width=2cm]{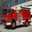}};
        
        \node at (2.5, 1.1) {\includegraphics[width=2cm]{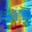}};
        \node at (2.5, -1.1) {\includegraphics[width=2cm]{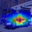}};

        \node at (4.8, 1.1) {\includegraphics[width=2cm]{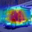}};
        \node at (4.8, -1.1) {\includegraphics[width=2cm]{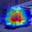}};

        \node at (7.1, 1.1) {\includegraphics[width=2cm]{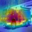}};
        \node at (7.1, -1.1) {\includegraphics[width=2cm]{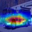}};

        \node at (0,-4.5) {\includegraphics[width=2cm]{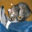}};
        
        \node at (2.5, -3.4) {\includegraphics[width=2cm]{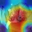}};
        \node at (2.5, -5.6) {\includegraphics[width=2cm]{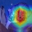}};

        \node at (4.8, -3.4) {\includegraphics[width=2cm]{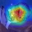}};
        \node at (4.8, -5.6) {\includegraphics[width=2cm]{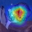}};

        \node at (7.1, -3.4) {\includegraphics[width=2cm]{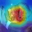}};
        \node at (7.1, -5.6) {\includegraphics[width=2cm]{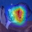}};

        \draw (-2,-6.75) -- (10.5,-6.75);
        \node[rotate=90] at (-1.75, -11.25) {\large\textbf{Imagenet-S}$_{\mathbf{50}}$};

        \node at (0,-9) {\includegraphics[width=2cm]{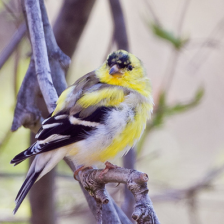}};
        
        \node at (2.5, -7.9) {\includegraphics[width=2cm]{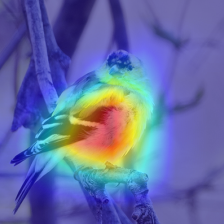}};
        \node at (2.5, -10.1) {\includegraphics[width=2cm]{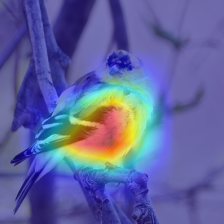}};

        \node at (4.8, -7.9) {\includegraphics[width=2cm]{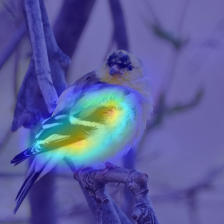}};
        \node at (4.8, -10.1) {\includegraphics[width=2cm]{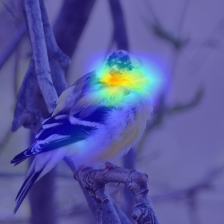}};

        \node at (7.1, -7.9) {\includegraphics[width=2cm]{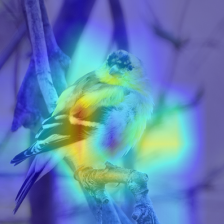}};
        \node at (7.1, -10.1) {\includegraphics[width=2cm]{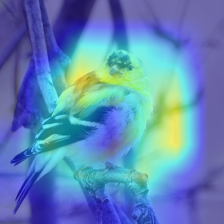}};

        \node at (0,-13.5) {\includegraphics[width=2cm]{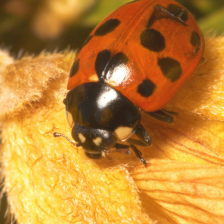}};
        
        \node at (2.5, -12.4) {\includegraphics[width=2cm]{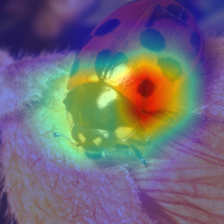}};
        \node at (2.5, -14.6) {\includegraphics[width=2cm]{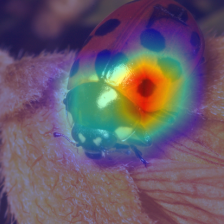}};

        \node at (4.8, -12.4) {\includegraphics[width=2cm]{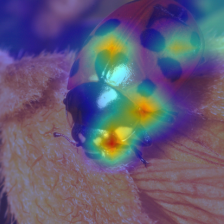}};
        \node at (4.8, -14.6) {\includegraphics[width=2cm]{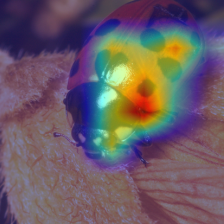}};

        \node at (7.1, -12.4) {\includegraphics[width=2cm]{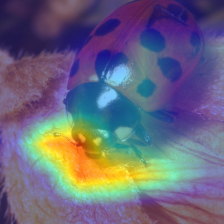}};
        \node at (7.1, -14.6) {\includegraphics[width=2cm]{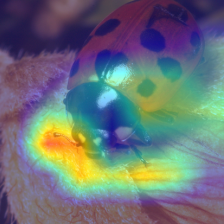}};

    \end{tikzpicture}
    }
    \caption{Sample explanations generated on CIFAR10 and Imagenet-S$_{50}$.
    A quick qualitative analysis shows how Grad-CAM for \ac{ce}, on CIFAR10, fails to focus on the object in the center of the image, rather producing wider \acp{fa}.
    This behavior seems to be present with less intensity on Imagenet-S$_{50}$.
    On both datasets, \ac{scl} seems to produce more compact explanations, which is confirmed by the lower complexity scores (see \Cref{tab:metrics}).
    The \ac{tl}-trained models seem to produce more erratic explanations in the case of Imagenet-S$_{50}$.
    Eigen-CAM seems to consistently produce compact \acp{fa}; however, its explanations cannot be generated for single classes, thus giving Grad-CAM an edge in that regard.}
    \label{fig:explanations}
\end{figure}

In \Cref{tab:accuracy}, we present the results for what concerns the accuracy of our ResNet-18 and ResNet-50 models trained using \ac{ce}, \ac{scl}, and \ac{tl}.
As explained in \Cref{sec:methods_models}, the accuracy for \ac{scl} and \ac{tl} is computed using a linear probe, i.e., the extra $f_\text{head}$ trained separately on the embedding space.
The accuracy scores highlight how all models achieve a good accuracy score, albeit, especially in Imagenet-S$_{50}$, the two \ac{cl} models showcase a substantially lower score than the \ac{ce} baseline.

\Cref{tab:xai_metrics} showcases the results with regard to the various metrics used to evaluate the explanations.
Despite the general trend, indicating that \acp{fa} generated from the \ac{scl}-trained models show better results across several metrics, there are two considerations to make:
\begin{enumerate*}[label=(\alph*)]
    \item \ac{tl} seems to lead to overall high-quality explanations on CIFAR10, while their evaluations on Imagenet-S$_{50}$ are often worse than \ac{scl}, and
    \item the coherence scores, computed only on Imagenet-S$_{50}$, for the availability of segmentation maps, are very noisy and do not allow us to determine a best method across the board.
\end{enumerate*}
All in all, the indication is that \ac{scl} seems to lead to better overall explanations.
For what concerns the specific explanation method is concerned, Grad-CAM seems to overall behave better than Eigen-CAM across several metrics.
This furthers the advantages of the former over the latter, since Grad-CAM \acp{fa} can be generated for multiple classes, while Eigen-CAM \acp{fa} is fixed and cannot be \emph{class-steered}. 

Finally, in \Cref{fig:explanations}, we show a small sample of explanations from the two datasets to better frame the results from \Cref{tab:xai_metrics}.
We can immediately notice how the explanations from \ac{tl} are substantially more focused on CIFAR10, while their equivalent on Imagenet-S$_{50}$ seem discontinuous and scattered, rarely focusing on one specific detail.
This can motivate the poor quality of explanations from \ac{tl} on the latter dataset.
Using Grad-CAM as a reference, we can also notice how the \acp{fa} seem more compact and focused on \ac{scl} than they are on \ac{ce}, which furthers the argument for \ac{scl} as a better choice for increasing transparency.

\section{Discussion and Conclusion}\acresetall

In the present work, we focused on comparing the quality of \ac{fa} explanations generated on models trained using two different supervised learning paradigms: the classical paradigm based on the \ac{ce} loss, and two supervised \ac{cl} approaches, one based on the \ac{scl}, the other on the \ac{tl}.
We trained a ResNet-18 on CIFAR10 and a ResNet-50 on Imagenet-S$_{50}$, each time with the three aforementioned loss functions.
All models reached good accuracy levels.
We then generated the \acp{fa} using the popular tools Grad-CAM and Eigen-CAM.
We then proceeded to evaluate the quality of the explanations on five facets: faithfulness/correctness, coherence, continuity, contrastivity, and complexity, as termed in the framework from Nauta et al.~\cite{nauta2023anecdotal}.
Our results indicate that \ac{scl}-trained models seem to produce overall better examples across both datasets, with the exception of contrastivity.
The other \ac{cl}-based loss, \ac{tl}, instead does not show promising results---despite good results on CIFAR10, the explanations on Imagenet-S$_{50}$ often seem scattered and unfocused, leading to poor results across all quality metrics.
With what concerns faithfulness, our results go in the direction of corroborating the previous findings connected to better \ac{ood} detection capabilities by \ac{cl}-trained models \cite{van2020uncertainty,zhou2021contrastive}: by being better \ac{ood} detectors, these models can also produce more reliable predictions when evaluated on the perturbed data produced by, e.g., \ac{pf}.
In this sense, it is not necessarily the case that explanations produced by \ac{ce}-trained models are less faithful, but rather that faithfulness can be computed more reliably by metrics such as \ac{pf}, without necessarily fine-tuning the models with data augmentation, such as in the case of F-Fidelity \cite{zheng2025ffidelity}.
As far as coherence is concerned, the results are too noisy to gather any conclusions.

Despite the positive findings, the present study has several limitations: first of all, we only ran our experiments on image classification tasks. 
This limitation is transversal to the field of \ac{xai} evaluation: image classification has an abundance of annotated datasets that can be leveraged for the computation of \ac{fa} properties, which are otherwise hard to assess (e.g., coherence), and the scale of these tasks makes the computational cost of the experiments not excessive.
{Secondly, it is to be noticed that our models across training objective have good but not equal accuracy values, which may consequently impact the quality of the explanation.}
Secondly, we could have included more metrics for \ac{xai} evaluation: faithfulness computation has different approaches, like ROAR \cite{hooker2019benchmark} and F-Fidelity \cite{zheng2025ffidelity} that could have been used; however, the first one 
\begin{enumerate*} [label=(\alph*)]
    \item focuses on the model's performance (instead of prediction), which Jacovi and Goldberg \cite{jacovi2020towards} suggest not being a proper evaluation of faithfulness, and
    \item requires retraining the model on the pixel-removed images,
\end{enumerate*}
while the second one requires additional fine-tuning, which may not always be applicable in real-world scenarios.
Finally, there are more recent approaches for \ac{cl} that could have been used, like variational supervised \ac{cl} \cite{wang2025variational} and ProjNCE \cite{jeong2025generalizing}, which have both been published recently.

Going forward, we plan on incorporating more approaches for \ac{cl} and widening the scope of our experiments and the selected metrics; additionally, we plan on further exploring the possibilities for generating explanations within the embeddings produced by the backbone of the models, without necessarily attaching an extra classification layer.
{
Additionally, we plan on running further analyses on the phenomenon of neural collapse \cite{papyan2020prevalence} and whether this may be responsible for the higher quality of the \acp{fa} in the case of \ac{scl}.}

All in all, the present paper gives positive indications that \ac{cl}-based approaches allow for the production of \acp{fa} with more desirable properties than \ac{ce}-based training, especially with regard to the complexity and coherence of the explanations.
We hope our findings can help guide practitioners in selecting training objectives that not only maximize accuracy but also facilitate the generation of high-quality explanations, ultimately fostering the development of more reliable and accountable artificial intelligence systems.

\bibliographystyle{splncs04} 
\bibliography{references}  

\end{document}